\documentclass{article}

\usepackage[accepted]{icml2020}

\icmltitlerunning{Low-loss connection of weight vectors: distribution-based approaches}

\usepackage{amsmath,amsfonts,bm}

\def\eqref#1{equation~\ref{#1}}

\def\1{\bm{1}}

\DeclareMathAlphabet{\mathsfit}{\encodingdefault}{\sfdefault}{m}{sl}
\SetMathAlphabet{\mathsfit}{bold}{\encodingdefault}{\sfdefault}{bx}{n}

\usepackage{microtype}
\usepackage{graphicx}
\usepackage{multirow}
\usepackage{subfigure}
\usepackage{booktabs} 
\usepackage{adjustbox}
\usepackage{hyperref}

\usepackage{url}
\usepackage{caption} 
\usepackage{longtable}
\usepackage{comment}
\usepackage{algorithmic}
\usepackage{algorithm}
\usepackage{todonotes}

\usepackage{tikz,tikz-cd,calc}
\usetikzlibrary{decorations.pathreplacing}
\usepackage{transparent}

\newtheorem{prop}{Proposition}

\def \opac {0.1}
\tikzcdset{row sep/verytiny/.initial=-0.3cm}
\definecolor{opaque_red}{rgb}{1,0.9,0.9}
\definecolor{opaque_blue}{rgb}{0.9,0.9,1}

\begin{document}

%

%

\twocolumn[

\icmltitle{Low-loss connection of weight vectors: distribution-based approaches}




\begin{icmlauthorlist}
\icmlauthor{Ivan Anokhin}{sk}
\icmlauthor{Dmitry Yarotsky}{sk}
\end{icmlauthorlist}

\icmlaffiliation{sk}{Skolkovo Institute of Science and Technology, Moscow}

\icmlcorrespondingauthor{Ivan Anokhin}{i.anokhin.mm@gmail.com}
\icmlcorrespondingauthor{Dmitry Yarotsky}{d.yarotsky@skoltech.ru}

\icmlkeywords{Machine Learning, ICML}

\vskip 0.3in
]



\printAffiliationsAndNotice{}  

\begin{abstract}
  Recent research shows that sublevel sets of the loss surfaces of overparameterized networks are connected, exactly or approximately. We describe and compare experimentally a panel of methods used to connect two low-loss points by a low-loss curve on this surface. Our methods vary in accuracy and complexity. Most of our methods are based on ``macroscopic'' distributional assumptions, and some are insensitive to the detailed properties of the points to be connected. Some methods require a prior training of a ``global connection model'' which can then be applied to any pair of points. The accuracy of the method generally correlates with its complexity and sensitivity to the endpoint detail.
\end{abstract}

\section{Introduction}

Though loss surfaces of neural networks have a complex shape, it is generally accepted that large networks train well and their performance is not very dependent on the weight initialization, despite apparently different trained values resulting from different initializations \cite{choromanska2015loss}. When thinking of the landscape of a complex nonconvex function such as a loss surface, one can imagine different heuristic scenarios for the structure of the bottom of the surface \cite{baity2018comparing}. One scenario is that the loss function has multiple isolated local minima. Another scenario is that there are, in contrast, few local minima, and the sub-level sets of the loss function have only one or a small number of connected components (despite their possibly complex shape). Of course, the second scenario agrees better with the practically observed efficiency of network training by gradient descent. In general, the second scenario is more likely in the setting of overparameterized networks (small networks are known to host numerous isolated local minima, see e.g. \cite{safran2017spurious}).  

Recent research provides some further evidence in favor of the ``connected sublevel set'' scenario. A particular easy-to-formulate task that one can analyze both experimentally and theoretically is: 
\begin{align}\label{connection_task}
    &\text{\it Given two low-loss weight vectors } \Theta^A, \Theta^B,\nonumber \\ 
    &\text{\it connect them by a low-loss curve.}
\end{align} Recent studies of this connectedness problem can be divided into numerical and theoretical ones. The numerical studies have been performed in \cite{garipov2018loss, draxler2018essentially}. In these papers, the desired low-loss curves are constructed by numerically optimizing the curves connecting the two given low-loss points. The results show that typically one can find a curve on which the loss value is only slightly worse than at the endpoints.

The theoretical studies rigorously confirm this effect under certain conditions (generally, overparameterization-related).  For a single-hidden- layer ReLU network, \cite{freeman2016topology} prove that two weight vectors with loss $\le l_0$ can be connected by a curve with loss $\le l_0$ if the number of hidden neurons is sufficiently large. For  pyramidal multilayer networks with piecewise linear activation functions, \cite{nguyen2019connected} proves that in the overparameterized setting (when the size of the first hidden layer is larger than the size of the training set), sublevel sets are connected and unbounded. \cite{kuditipudi2019explaining} assume that the model is dropout-stable or noise-stable, and then construct a connecting path with a low loss.  

Obviously, these experimental and theoretical works have quite different methodologies. The 
paths found in the experimental studies are numerically optimized to particular endpoints, and the structure of these optimal paths is not well understood. On the other hand, while the theoretical works offer some explicit rigorous constructions of low-loss paths, it is not clear to which extent they match the experimentally found ones.    

Motivated by this discrepancy, in the present paper we adopt a somewhat different point of view on task (\ref{connection_task}),  putting forward this general goal:

\emph{Describe universally applicable and possibly simple  methods that, given two weight vectors $\Theta^A, \Theta^B$ produce connecting curves of possibly low loss.}

With this goal in mind, we propose a panel of methods of different complexity and accuracy,  bridging the gap between the above numerical and theoretical studies. 

In contrast to the numerical optimization of \cite{garipov2018loss, draxler2018essentially}, we aim to construct the connecting curve by a more-or-less direct prescription (to a varying degree, depending on the method). While the above numeric optimization papers demonstrate but do not explain the connectedness phenomenon, our methods logically follow from either the particular form or certain assumptions about the trained networks. 

On the other hand, in contrast to the mentioned theoretical studies, we are interested in ``general-purpose'' low-loss connection methods that are, in principle, applicable to any pair of endpoints $\Theta^A,\Theta^B$, any network size, and any training data. 
To clarify this point, consider the most trivial connection performed by a straight line segment: $t\mapsto \Theta(t)=(1-t)\Theta^A+t\Theta^B$. The performance of this method is quite poor (the loss can grow significantly for intermediate $t$), but the method is universally applicable, given by an explicit analytic formula, and the geometry of the path is essentially independent of the values $\Theta^A,\Theta^B.$  The papers \cite{freeman2016topology, nguyen2019connected} represent another extreme case, where $\Theta^A$ and $\Theta^B$ are connected using a complex path and meticulous adjustment of individual neuron parameters (and only under rather restrictive assumptions on the model), but achieving a perfect solution of task (\ref{connection_task}). In the present paper, we are interested in the intermediate setting: possibly generally applicable, endpoint-insensitive methods with possibly simple paths, yet improving performance of the trivial straight-line connection. 

\section{Our contribution and the structure of the paper}
We start by considering networks with a single hidden layer (Section \ref{sec:shallow}). Our connection methods are largely motivated by the ``macroscopic'' view of the network as a sample from some distribution in a suitable state space of neurons; we recall this picture in Section \ref{sec: loss_2_dist}.  
\begin{itemize}
    \item In Section \ref{sec: methods} we describe the idea of connections preserving the neuron distribution, and specifically describe ``Arc Connection'' which is an analytic method essentially as simple as the trivial segment connection, but preserving the variance of the neuron distribution. The Arc Connection is a perfect solution of the connection problem in the limit of infinitely wide networks if the neurons are normally distributed.
    \item In Section \ref{sec: learnable_methods} we generalize Arc Connection to non-Gaussian distributions of neurons. To this end, we introduce ``learnable methods'' aimed to learn the neuron distribution in a typical local minimum of the loss function. In this way, we construct a ``global connection model'' that can be used subsequently to connect any two new local minima.     
    \item In Section \ref{sec:ot} we describe ``Optimal Transportation'' methods in which the connecting path consists of two stages. In the first stage the distribution of neurons in one local minimum is optimally transported to the distribution in another minimum, and in the second stage the neurons are permuted to be in the required order.
    \item Finally, in Section \ref{sec:weightadjust} we describe what we call ``Weight Adjustment'' methods in which the first layer weights are connected by a simple analytic prescription while the second layer weights (on which the network output depends linearly) are adjusted appropriately, by solving suitable linear systems.  
\end{itemize}
In Section \ref{sec:ext} we extend these methods to multi-layer networks (by a suitable layer-wise reduction) and convnets.

In Section \ref{sec: Experiments} we perform an experimental comparison of these connection methods. 

Finally, in Section \ref{sec:ensembling} we discuss one potential practical application of the connection task: one can use low-loss connections between different low-loss weight vectors to form an ``ensemble'' of networks with an accuracy slightly better than that of the individual networks. In contrast to conventional ensembles, this can be achieved with only a small computational overhead on inference, by reusing initial computation of one of the networks. 

\section{One Hidden Layer}\label{sec:shallow}
\subsection{Reduction to Distributions}\label{sec: loss_2_dist}
The theory of networks with a single hidden layer can be relatively easily translated into the language of distributions, so that the network output, the loss function, and the gradient descent are described in terms of the weight distributions rather than individual values. We sketch the main ideas, referring the reader to the papers \cite{mei2018mean, rotskoff2018neural, sirignano2018mean, chizat2018global} for details and precise statements.  

\begin{figure}[h]
\centering
\includegraphics[width=.7 \linewidth]{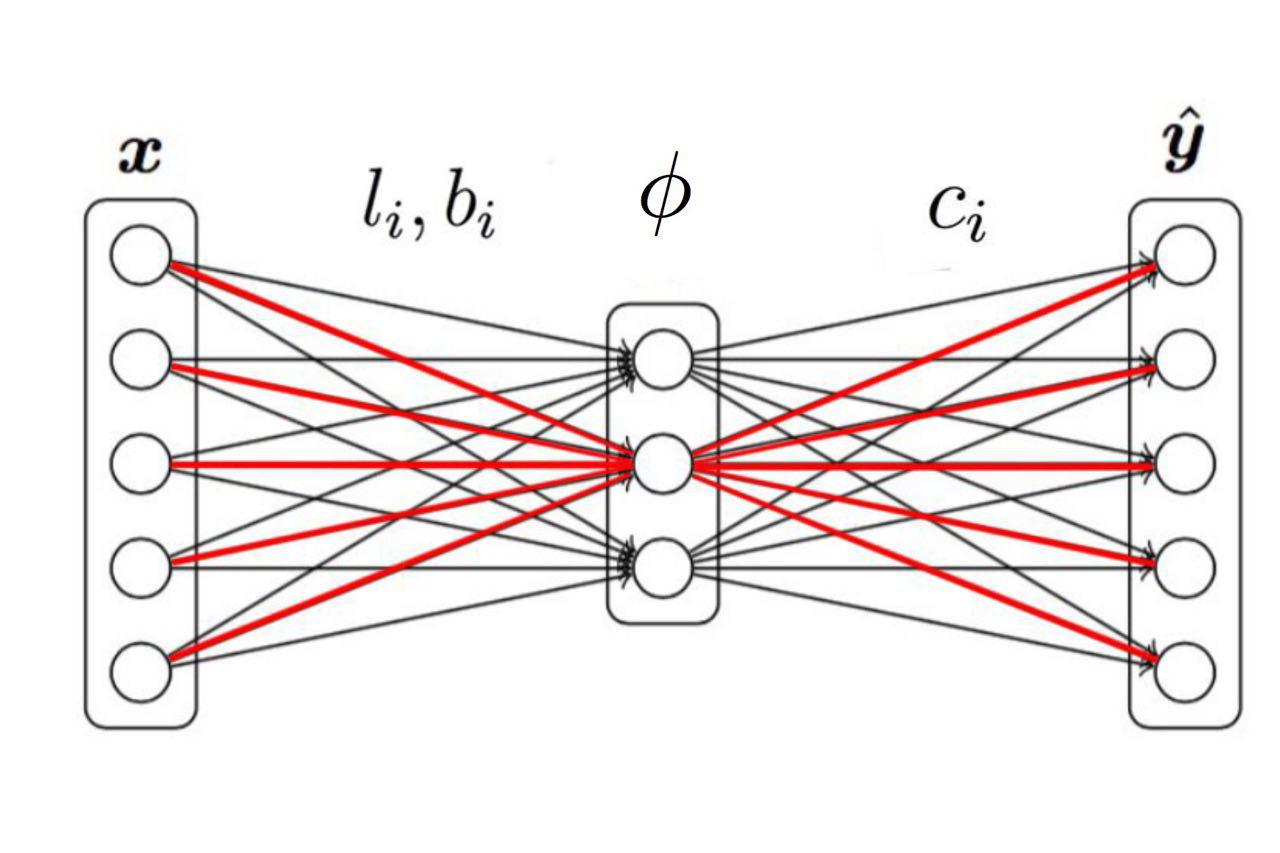}
 \caption{A graphical representation of a one hidden layer network, as in Eq.~(\ref{onelayer_sum}). Weights of one particle $\mathbf \theta_i = (b_i, \mathbf l_i, \mathbf c_i)$, colored in red, resemble a butterfly.}
 \label{butterfly}
\end{figure}

Consider the predictive model of the form
\begin{equation}
\label{onelayer_sum}
\mathbf{\widehat{y}}_n (\mathbf{x}; \Theta) = \frac 1 n \sum_{i=1}^n \sigma (\mathbf{x}; \mathbf \theta_i ) ,
\end{equation}
where $\mathbf{x} \in \mathbb{R}^d$ is the input, $\Theta = \{\mathbf \theta_i \}_{i \in [n]}$
is the collection of weights $\mathbf{\theta}_i \in \mathbb{R}^D$, 
and ${\sigma : \mathbb{R}^d \times \mathbb{R}^D \rightarrow \mathbb{R}^m}$ is some map.
In particular, we obtain the standard fully-connected neural network with a single hidden layer by setting $\mathbf \theta_i = (b_i
, \mathbf l_i, \mathbf c_i ) \in \mathbb{R} \times \mathbb{R}^d \times \mathbb{R}^m$
and $\sigma (x; \mathbf \theta_i) = \mathbf c_i \phi(\langle \mathbf l_i
,x_i \rangle + b_i)$ with a scalar activation function $\phi$. Each term in the sum then corresponds to a hidden neuron, see Fig.~\ref{butterfly}. 

Let us now write predictive model (\ref{onelayer_sum}) in the form
\begin{equation}
\label{onelayer_int}
    \mathbf{\widehat{y}}(\mathbf{x}; p) = \int \sigma (\mathbf{x}; \theta) p(d \theta),
\end{equation}
where $p$ is the normalized counting measure on the space $\mathbb R^D$ concentrated at the weights $\mathbf{\theta}_i$: $p= \frac 1 n \sum_{i=1}^n \delta_{\mathbf{\theta}_i}$. One important advantage of this new representation in terms of $p$ is that we get rid of the excessive degree of freedom associated with permutations of neurons. Another advantage is that formula (\ref{onelayer_int}) naturally generalizes to any measure $p$ on $\mathbb R^D.$ Finally, while the representation (\ref{onelayer_sum}) is, in general, not linear in the weights $\mathbf{\theta}_i$, the representation (\ref{onelayer_int}) is linear in $p$, so that if the loss function is convex in $\mathbf{\widehat{y}}$, it is also convex in $p$. In the sequel, we will assume this convexity of the loss.    

The gradient descent for the model (\ref{onelayer_sum}) can also be described in terms of $p$, by a suitable integro-differential equation,
and the dependence of the GD trajectory on the initial distribution is sufficiently regular. This suggests the following approach to the connection task (\ref{connection_task}). Suppose that the weight vectors $\Theta^A,\Theta^B\in\mathbb R^D$ have been obtained by optimizing the loss function starting from two different random initializations $\Theta_0^A, \Theta_0^B$ obtained  by sampling the weights independently from the same initial distribution $p=p_0$ (for example, by sampling the weights as i.i.d. normal variables, as is the usual practice). Then, by the law of large numbers, for a large network we expect the initial distributions $p_0^A=\tfrac{1}{n}\sum_{i=1}^n \delta_{\mathbf{\theta}^A_{0,i}}, p_0^B=\tfrac{1}{n}\sum_{i=1}^n \delta_{\mathbf{\theta}^B_{0,i}}$ to be close to $p_0$, and hence expect their whole gradient descent trajectories to be close in the space of distributions. Accordingly, the final optimized weight vectors $\Theta^A, \Theta^B$ should also be described by approximately the same distribution $p$. Then, to connect the points $\Theta^A$ and $\Theta^B$ by a low-loss path $\psi:t\in[0,1]\mapsto\Theta(t)\in\mathbb R^D$, we want to choose it in such a way that the weight distribution for $\Theta(t)$ is also approximately equal to $p$ for all $t$.  If we manage to do so, the output of neural networks along this path will be approximately the same:
\begin{equation}
\mathbf{\widehat{y}}_n (\mathbf{x}; \psi(t)) \approx  \mathbf{\widehat{y}}_n (\mathbf{x}; p).    
\end{equation}

\subsection{Distribution Preserving Methods}
\label{sec: methods}

The above arguments suggest reducing the connection task (\ref{connection_task}) to constructing a ``distribution-preserving'' deformation. Specifically, let $X$ and $Y$ be two independent random vectors of length $D$ sampled from an unknown distribution $p$ on $\mathbb R^D$. We want to construct a continuous path $\psi : [0, 1] \rightarrow \mathbb{R}^D$ such that $\psi(0)=X, \psi(1)=Y$, and the distribution of the random vector $\psi(t)$ is $p$ for any $t \in [0,1]$. Once we have such a method, we can apply it to connect the network weight vectors $\Theta^A,\Theta^B$ in a component-wise way:
$$\Theta(t)=(\psi_i(t))_{i=1}^n,$$
where $\psi_i$ connects $X=\mathbf{\theta}_i^A$ to $Y=\mathbf{\theta}_i^B.$

Now we will consider several particular methods of connecting $X$ to $Y$, and we start from the trivial baseline:

\textbf{Linear Connection} is  the basic most naive way to connect two weight vectors: 
\begin{equation}
\label{eq:linconnect}
\psi(t) = (1-t)X + tY.
\end{equation}
Note that this method is not measure-preserving, in general: if $X,Y\sim p,$ then $\psi(t)\not\sim p$ for $t\in(0,1)$. This can be seen, for example, by considering the covariance matrix $\Sigma_{\psi(t)}$ of $\psi(t)$, which is equal to $(1-t)^2\Sigma_X+t^2\Sigma_Y=((1-t)^2+t^2)\Sigma_p\ne\Sigma_p$ (so the Linear Connection ``squeezes'' the distribution $p$). This explains why performance of the Linear Connection is typically rather poor. 

\begin{figure}[h]
\centering
{\includegraphics[width=.8 \linewidth, clip, trim=30mm 17mm 30mm 30mm]{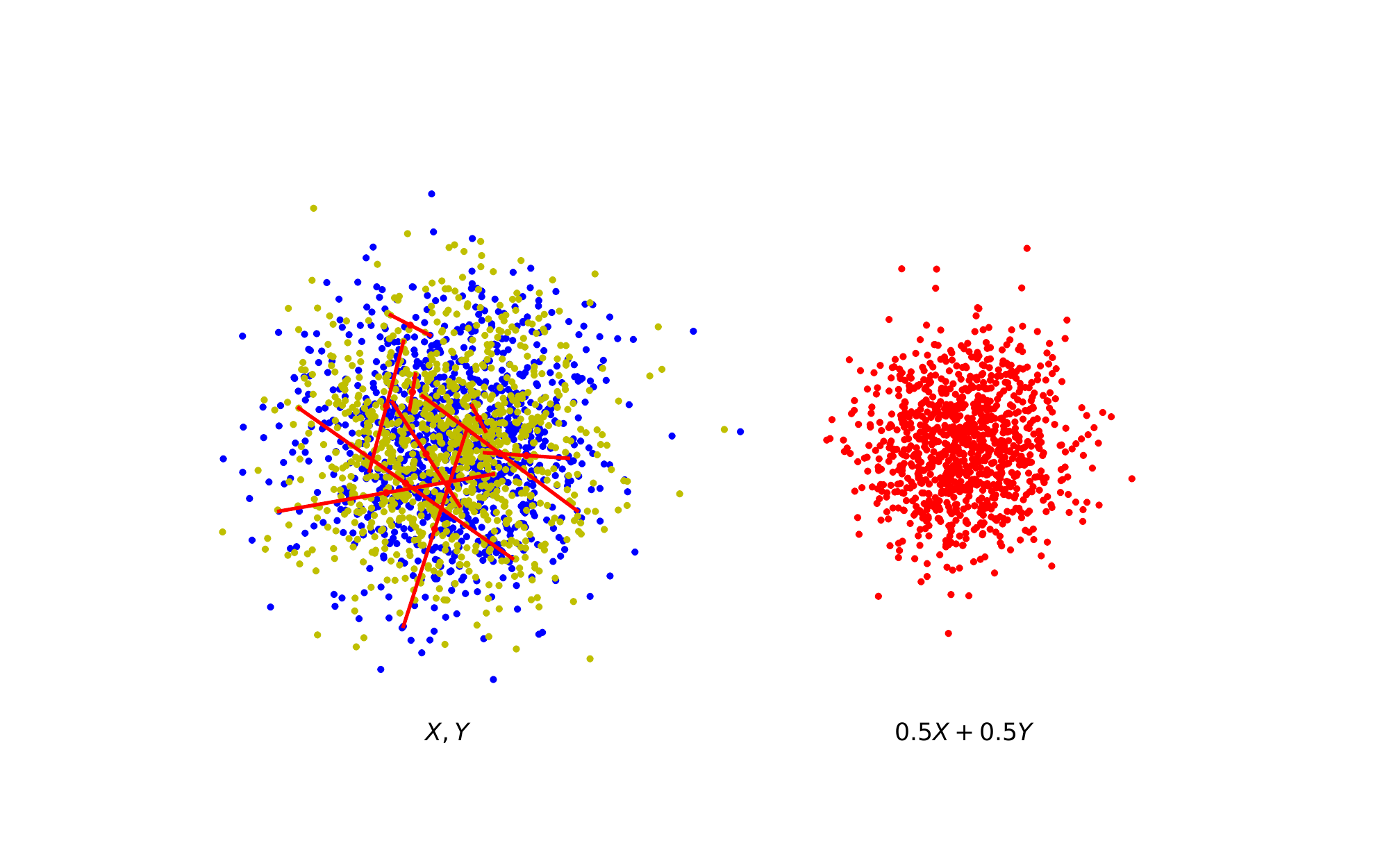}}
{\includegraphics[width=.8 \linewidth, clip, trim=30mm 17mm 30mm 30mm]{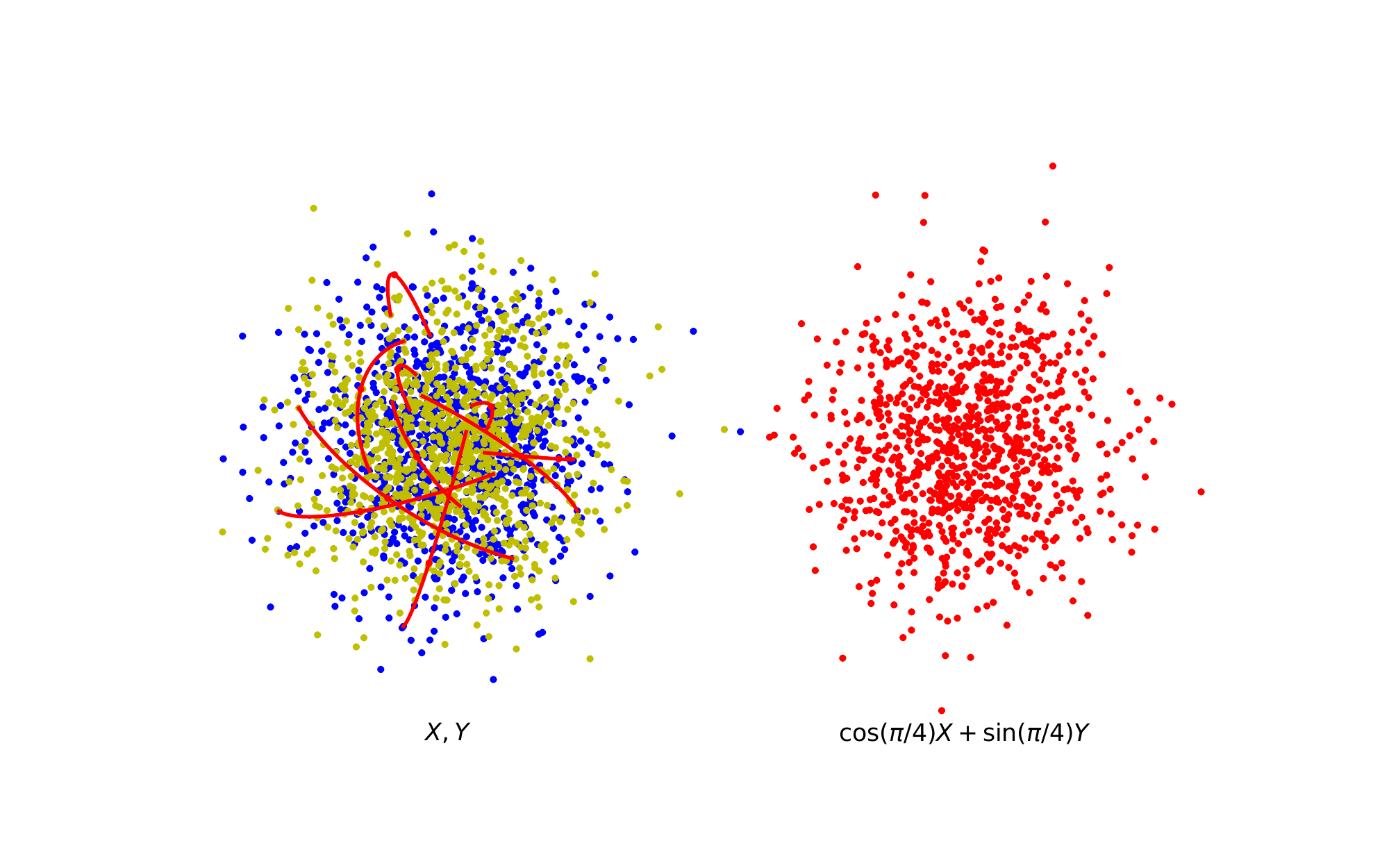}}
 \caption{Connection of two samples $X,Y\sim\mathcal N(0,\mathbf 1_{2\times 2})$. \textbf{Top:} Linear, Eq.(\ref{eq:linconnect}), squeezes the distribution. \textbf{Bottom:} Arc, Eq.(\ref{normal_transform}), preserves the distribution.}
 \label{fig:arcConnect}
\end{figure}

The following proposition suggests how to modify formula (\ref{eq:linconnect}) to make the connection measure preserving in the case of a multivariate Gaussian distribution $p$  (see Fig.~\ref{fig:arcConnect}).
\begin{prop}
\label{gauss_prop}
If $X, Y$ are i.i.d. vectors with the same centered multivariate Gaussian distribution $p$, then for any $t \in \mathbb{R}$, ${\psi(t) = \cos(\frac \pi 2 t)X + \sin(\frac \pi 2 t)Y}$ has the same distribution  $p$, and also $\psi(0) = X, \psi(1) = Y$.
\end{prop}
One can easily prove this known fact by using characteristic function of multivariate normal distribution.
We can then give our first improvement of Linear Connection.

\textbf{Arc Connection} is the method that assumes that $X, Y$ are already Gaussian with the same covariance matrix and center: $\mu = \mathbb{E}X = \mathbb{E}Y$. Then, we set: 
\begin{align}
\label{normal_transform}
\psi(t) = {}&\mu+ \cos(\tfrac \pi 2 t)(X-\mu) + \sin(\tfrac \pi 2 t)(Y-\mu).
\end{align}

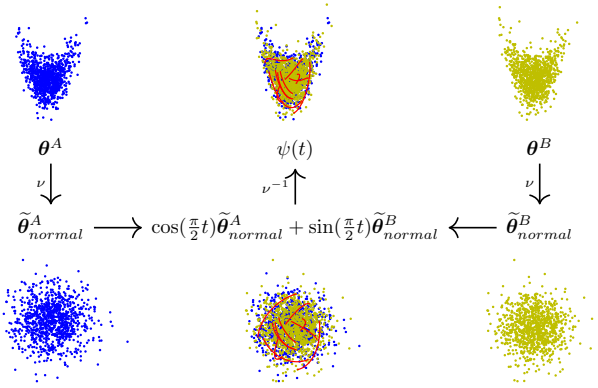
\begin{figure}
\begin{center}
\includegraphics[width=1\linewidth, trim={40mm 5mm 35mm 10mm}, clip]{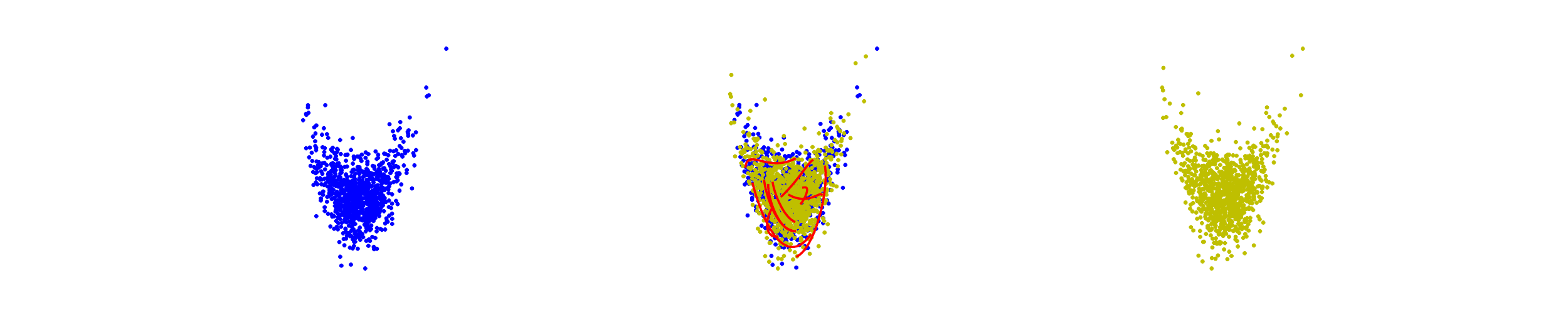}

\adjustbox{scale=0.77,center}{
\begin{tikzcd}
\boldsymbol{\theta}^A
\arrow[d, "\nu"', thick]
& 
\psi(t) 
&
\boldsymbol{\theta}^B
\arrow[d, "\nu"', thick]\\ 
\widetilde{\boldsymbol{\theta}}^A_{normal}\arrow[r, thick]& \cos(\frac \pi 2 t)\widetilde{\boldsymbol{\theta}}^A_{normal} + \sin(\frac \pi 2 t)\widetilde{\boldsymbol{\theta}}^B_{normal}\arrow[u, "{\nu^{-1}}", thick]&\widetilde{\boldsymbol{\theta}}^B_{normal}\arrow[l, thick
]
\end{tikzcd}
}


\includegraphics[width=1\linewidth, trim={37mm 5mm 35mm 5mm}, clip]{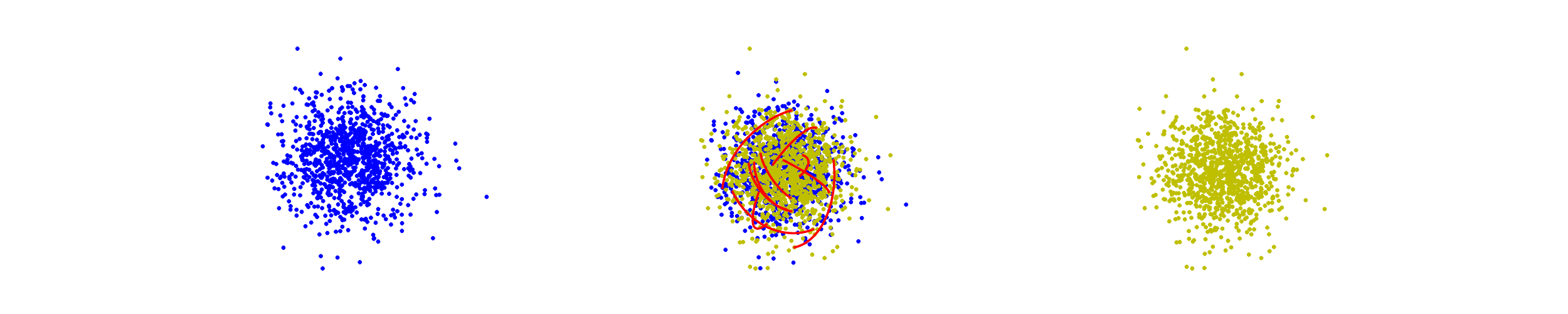}
\end{center}
    \caption{
    General distribution-preserving path, Eq.(\ref{main_transform}), 
    maps target distributions $\boldsymbol{\theta}^A, \boldsymbol{\theta}^B$ to standard normal, mixes them, and maps the mix back.}
    \label{fig:nu}
\end{figure}

However, the assumed normality is a severe restriction: the distribution of weights is non-normal in general. We can generalize the Arc Connection to non-normal $X,Y$ by considering a general transformation $\nu$ making $X,Y$ normal:  
\begin{equation}
\label{main_transform}
{\psi(t) = \nu^{-1}[\cos(\tfrac \pi 2 t)\nu(X) + \sin(\tfrac \pi 2 t)\nu(Y)]}
\end{equation}
(see Fig.~\ref{fig:nu}). In practice, we don't know the map $\nu$, but we can try to learn a suitable map from the data. This leads us to what we refer to as learnable connection methods. 


\subsection{Learnable Connection Methods}
\label{sec: learnable_methods}

We want to learn a suitable transformation $\nu$ in Eq.~(\ref{main_transform}). For this purpose we propose to use neural network architectures that support inverse transformation (note that Eq.~(\ref{main_transform}) requires us to compute both $\nu$ and its inverse). 
Such architectures are often used in normalizing flows methods, which aim to transform  simple known probability distribution (e.g. Gaussian) into a complicated multi-modal one and still be able to compute the probability of the point.  The training and fast inference of the models are achieved by using transformations whose Jacobian determinants are easy to compute.  

Our learnable methods are characterized by two elements: the network architecture used to compute $\nu$, and the optimization algorithm used to train the network. We utilize two architectures as $\nu$. The first one is the \textbf{RealNVP} model as described in \cite{dinh2016density}, the second one is the Inverse Autoregressive Flow (\textbf{IAF}) model as described in \cite{kingma2016improved}. 
To emphasize that now transformation $\nu$ has parameters, we  write it as $\nu_W$, where $W$ are the weights of one of the above two networks.

We consider two optimizing procedures, \textbf{Flow} and \textbf{Bijection}.
In \textbf{Flow}, we follow the algorithm proposed in  \cite{dinh2016density} and maximize the likelihood $\mathbb E_{\mathbf x\sim p}\log\big[\eta(\nu(\mathbf x)) |\det \tfrac {\partial \nu( \mathbf x)} {\partial \mathbf x}|\big]$, 
where $\eta$ is the standard normal probability density function. 
After training is done, we can use Eq.~(\ref{main_transform}) to generate samples along the path. Note that the transformation $\nu$ should map samples from the target distribution to the standard Gaussian if the training procedure is successful. 

We also propose a new training procedure which we call \textbf{Bijection}. 
Assume we have a dataset $V=\{\Theta\}$ of low loss weight vectors for a One Hidden layer network.  We can easily create such $V$ by training models that  minimize any user-specified loss $L(\Theta)$.
Now we want to have low loss for any two models in $V$ and any point $t$ on the curve (\ref{main_transform}) that is convenient to rewrite as 
\begin{align*}
    \psi_W & (t, \Theta^A, \Theta^B) \\
    & = \nu_W^{-1}[\cos(\tfrac \pi 2 t)\nu_W(\Theta^A) +  \sin(\tfrac \pi 2 t)\nu_W(\Theta^B)].
\end{align*}
Similarly to \cite{garipov2018loss}, in order to achieve this  we propose to optimise the computationally  tractable loss 
\begin{equation}
\label{bijection_loss}
l(W) = \mathbb{E}_{*} L(\psi_W(t, \Theta^A, \Theta^B)),
\end{equation}
where expectation $\mathbb{E}_*$ is w.r.t. $t \sim U(0,1), \Theta^A \sim U(V), \Theta^B \sim U(V \setminus \Theta^A).$
To minimize Eq.~ (\ref{bijection_loss}), at each iteration we sample $\widehat{t}$ from the uniform distribution $U(0, 1)$,  $\widehat\Theta^A, \widehat\Theta^B$ are drawn from $V$ uniformly in a way that $\widehat\Theta^A \ne \widehat\Theta^B$, then we make a gradient step for $W$ with respect to the loss $L(\psi_W(\widehat t, \widehat \Theta^A, \widehat \Theta^B))$.
We repeat these updates until convergence.

We have found experimentally that it is usually sufficient to optimize the model only in the middle point $t=0.5$, as the model tends to always have the highest loss there:
\begin{equation*}
\label{bijection_loss_middle}
l(W) = \mathbb{E}_{\Theta^A \sim U(V), \Theta^B \sim U(V \setminus \Theta^A)} L(\psi_W(0.5, \Theta^A, \Theta^B)).
\end{equation*}
Strictly speaking, \textbf{Bijection}-based connection methods are not constructed as distribution-preserving along the path, but they are expected to generate low-loss paths between any similarly trained models. However, let us note that one possible solution for  \textbf{Bijection} procedure is to learn a map to centered Gaussian distribution. 

We name learnable methods in the following manner: the first part of the name is a network architecture name, and the second is a training procedure name. Combining various architecture and training procedure, we get four connection methods: \textbf{RNVP Flow}, \textbf{IAF Flow}, \textbf{RNVP Bijection} and \textbf{IAF Bijection}. However, training a network with \textbf{Bijection} requires a fast computation of $\nu_W$ and $\nu_W^{-1}$. For this reason, in this case we use only \textbf{RealNVP} networks, not \textbf{IAF}. 

Note that the approach proposed in this section can be described as ``training a global connection model''. We perform a single initial training of this model, but once done, we can connect any pair of unseen samples from the distribution $p$ using our learned transformation $\nu_W$. In terms  of connecting network weights, this means that we can use this global model to connect any pair of weight vectors, assuming they have been trained in the way similar to the one used to generate the training data for the global connection model.

\subsection{Optimal Transportation Methods}\label{sec:ot}
We consider now an alternative approach (referred to as \textbf{OT} in the sequel) that is also based on the idea of connecting two distributions, but attempts to do it by taking into account the whole set of ``butterflies''. Specifically, we use a version of Optimal Transportation (OT) in the neuron state space $\mathbb R^D$ to connect the sample of hidden neurons of the network $A$ to that of $B$. If the number of neurons is large, then we can find a bijective map between the neurons of $A$ and $B$ that maps each neuron of $A$ to a nearby neuron of $B$. In this way, we can transform the network $A$ to a network isomorphic to $B$ (namely, different from $B$ only by the order of hidden neurons) by a short linear segment in the full weight space $\mathbb R^{nD}$, so that the distribution of neurons remains approximately constant on this segment. We use the POT library \cite{flamary2017pot} for the solution of this OT problem. 

Note, however, that the OT transformation alone does not solve our connection task, since this task requires us to connect each neuron of $A$ to a particular target neuron of $B$, i.e., keep the prescribed order of neurons. Therefore, we supplement the above OT-stage of the path by the ``permutation'' stage. This second stage can be implemented by a continuous piecewise linear curve adjusting the neurons one-by-one. In each step, a pair of neurons is swapped placing one of them at the required position. The swap can be implemented by a singe linear transformation. Since the contribution of each hidden neuron to the network output is $O(1/n)$ and completed swaps do not change the network output, this path maintains low values of the loss function.

\subsection{Joint Weight Adjustment}
\label{sec:weightadjust}

For completeness, we also consider a connection method that goes beyond the distributional picture and uses a direct analytic weight adjustment for the given pair of weight vectors $\Theta^A,\Theta^B.$ Let us write a network with a single hidden layer in the standard form
$$\widehat{\mathbf{y}} = W_2\phi(W_1\mathbf x),$$
where $W_1,W_2$ are matrices (of size $d_1\times d_0, d_2\times d_1$, respectively), and the activation function $\phi$ is meant to act separately on each component of the vector $W_1\mathbf x$. For simplicity, we do not include the bias terms in this formula (the bias can be introduced in the first layer by assuming that $\mathbf x$ has an additional component with a constant value).

Let $\Theta^A= (W^A_1,W^A_2)$ and $\Theta^B=(W^B_1,W^B_2)$ be two weight vectors for which the network has close outputs. This condition can be written as follows. Let $\mathbf X$ be the $d_{0}\times N$ matrix of the set $S=\{\mathbf x_q\}_{q=1}^N\in\mathbb R^{d_0}$ of $N$ input vectors  on which we consider the action of the network. On this set $S,$ the network output can be written as
$$\widehat{\mathbf Y} = W_2\phi(W_1\mathbf X).$$
Let $\widehat{\mathbf Y}^A, \widehat{\mathbf Y}^B$ be the outputs with the weight values $\Theta^A,\Theta^B;$ we then assume that $\widehat{\mathbf Y}^A\approx \widehat{\mathbf Y}^B.$

We choose now a path $\Theta=\Theta(t)=(W_1(t),W_2(t)), t\in[0,1],$ that connects $\Theta^A$ to $\Theta^B$ and approximately preserves the output $\widehat{\mathbf Y}.$ To this end, we first connect  $W_1^A$ to $W_1^B$ in a more or less arbitrary way, for example using the basic linear connection $W_1(t)=(1-t)W^A_1+tW^B_1.$ Then, we adjust the weights in the second layer, which essentially means that we need to solve the linear system
\begin{equation}\label{eq:twolayerlineq}W_2(t)\phi(W_1(t)\mathbf X)\approx \widehat{\mathbf Y}^A\end{equation}
for $W_2(t)$, at each $t\in[0,1].$ A solution can be written as 
\begin{equation}\label{eq:twolayersol}W_2(t) = \widehat{\mathbf Y}^A \Big[\phi(W_1(t)\mathbf X)\Big]^+,\end{equation}
where $[\cdot]^+$ denotes the pseudo-inverse matrix. In general, this solution may be discontinuous in $t$ (at the points where the rank of $\phi(W_1(t)\mathbf X)$ changes), and the boundary values $W_2(0),W_2(1)$ may be different from $W_2^A,W_2^B$. The last issue is related to the degeneracy of the system (\ref{eq:twolayerlineq}) and, by linearity, can be resolved simply by adding to the path (\ref{eq:twolayersol}) two extra legs linearly connecting $W_2^A$ to $W_2(0)$ and $W_2(1)$ to $W_2^B.$ As for discontinuity, we resolve this issue by applying Eq.~(\ref{eq:twolayersol}) only at finitely many values of $t$, and forming the full path as the piecewise linear curve with these breakpoints.

Let us introduce the following name convention for variants of this weight adjustment method. The first part of the name refers to the distribution interpolation method that we use to connect the weight vectors $W_1^A$ and $W_1^B$. We connect them by considering the lines of the matrix $W_1$ as sampled from an unknown distribution $p$.  The second part of the name emphasizes that we use the weight adjustment in the second layer. In particular, we make experiments with the methods \textbf{Linear + Weight Adjustment}, \textbf{Arc + Weight Adjustment} and  \textbf{OT + Weight Adjustment}.  

\section{Extensions to more complex networks}\label{sec:ext}
\subsection{Multi-layer networks}
\label{sec:multi}

Let us introduce some additional notations:   $\mathbf X_k = \phi(W_{k}\phi(\ldots\phi(W_1\mathbf X)\ldots))$ is the input for the layer $k$,  $\mathbf X_0 = \mathbf X$ is the initial input for the network, $W_{k+1}^{AB} = W_{k+1}^A \mathbf X_{k}^A\Big[\mathbf X_{k}^B\Big]^+$ is the weight adjustments of the $k$'th layer of network $A$  to the $k$'th layer of network $B$ (as in Eq.~(\ref{eq:twolayersol})), and $\Theta_k^{AB} = \{W_1^B, ..., W_{k-1}^B, W_{k}^{AB}, W_{k+1}^A, W_{k+2}^A, ..., W_{n}^A\}$ are intermediate points that we cross on the way from the weights $\Theta^A$ to $\Theta^B$.

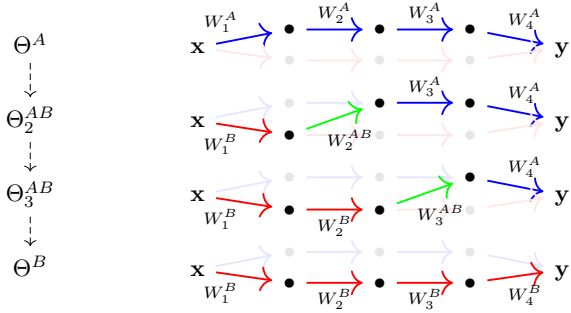
\begin{figure}
    \centering
\adjustbox{scale=0.9,center}{
\begin{tikzcd}[row sep=verytiny]
&&& 
{\transparent{1}\bullet}
\arrow[r, "W_2^A" black, opacity=1, thick, blue]
& 
{\transparent{1}\bullet} 
\arrow[r, "W_3^A" black, opacity=1,
blue, thick] 
& 
{\transparent{1}\bullet} 
\arrow[dr, "W_4^A" {black, near start}, opacity=1,
blue, thick] 
\\
\Theta^A \arrow[ddd, dashed]&&
\mathbf{x}
\arrow[ur, "W_1^A" black, opacity=1, blue, thick]
\arrow[dr, 
opacity=\opac, opaque_red, thick]
& & & &
\mathbf{y}
\\
&&& 
{\transparent{\opac}\bullet}
\arrow[r, 
opacity=\opac, thick, opaque_red] 
& 
{\transparent{\opac}\bullet}
\arrow[r, 
opacity=\opac, thick, opaque_red]
& 
{\transparent{\opac}\bullet}
\arrow[ur, 
opacity=\opac,
opaque_red, thick] 
\\[5mm]
&&& 
{\transparent{\opac}\bullet}
\arrow[r, 
opacity=\opac, thick, opaque_blue]
& 
{\transparent{1}\bullet} 
\arrow[r, "W_3^A" black, 
opacity=1,
blue, thick] 
& 
{\transparent{1}\bullet} 
\arrow[dr, "W_4^A" {black, near start}, opacity=1,
blue, thick] 
\\
\Theta_2^{AB} \arrow[ddd, dashed]&&
\mathbf{x}
\arrow[ur, 
opacity=\opac, opaque_blue, thick]
\arrow[dr, "W_1^B"' black, opacity=1, red, thick]
& & & &
\mathbf{y}
\\
&&& 
{\transparent{1}\bullet}
\arrow[r, 
opacity=\opac, thick, opaque_red] 
\arrow[uur, "W_2^{AB}"' {black, near start}, opacity=1, thick, green] 
& 
{\transparent{\opac}\bullet}
\arrow[r, 
opacity=\opac, thick, opaque_red]
& 
{\transparent{\opac}\bullet}
\arrow[ur, 
opacity=\opac, opaque_red, thick] 
\\[5mm]
&&& 
{\transparent{\opac}\bullet}
\arrow[r, 
opacity=\opac, thick, opaque_blue]
& 
{\transparent{\opac}\bullet} 
\arrow[r, 
opacity=\opac,
opaque_blue, thick] 
& 
{\transparent{1}\bullet} 
\arrow[dr, "W_4^A" {black, near start}, opacity=1,
blue, thick] 
\\
\Theta_3^{AB}\arrow[ddd, dashed] &&
\mathbf{x}
\arrow[ur, 
opacity=\opac, opaque_blue, thick]
\arrow[dr, "W_1^B"' black, opacity=1, red, thick]
& & & &
\mathbf{y}
\\
&&& 
{\transparent{1}\bullet}
\arrow[r, "W_2^B"' black, opacity=1, thick, red] 
& 
{\transparent{1}\bullet}
\arrow[r, 
opacity=\opac, thick, opaque_red]
\arrow[uur, "W_3^{AB}"' {black, near start}, opacity=1, thick, green] 
& 
{\transparent{\opac}\bullet}
\arrow[ur, 
opacity=\opac,
opaque_red, thick] 
\\[5mm]
&&& 
{\transparent{\opac}\bullet}
\arrow[r, 
opacity=\opac, thick, opaque_blue]
& 
{\transparent{\opac}\bullet} 
\arrow[r, 
opacity=\opac,
opaque_blue, thick] 
& 
{\transparent{\opac}\bullet} 
\arrow[dr, 
opacity=\opac,
opaque_blue, thick] 
\\
\Theta^B &&
\mathbf{x}
\arrow[ur, 
opacity=\opac, opaque_blue, thick]
\arrow[dr, "W_1^B"' black, opacity=1, red, thick]
& & & &
\mathbf{y}
\\
&&& 
{\transparent{1}\bullet}
\arrow[r, "W_2^B"' black, opacity=1, thick, red] 
& 
{\transparent{1}\bullet}
\arrow[r, "W_3^B"' black, opacity=1, thick, red]
& 
{\transparent{1}\bullet}
\arrow[ur, "W_4^B"' {black, near start}, opacity=1,
red, thick] 
\end{tikzcd}
}
\caption{
Intermediate points on the path from a four-layer network $A$ to a network $B$. Starting from the first layer, the weights of model $A$ are gradually replaced with weights of model $B$.
}
\label{fig: four-layer-nn-path}
\end{figure}

We propose to connect two weight vectors, $\Theta^A$ and $\Theta^B$, of a multi layer dense net   with the following intermediate points: $\Theta^A \to \Theta_2^{AB} \to  \Theta_3^{AB} \to ,..., \to \Theta_n^{AB} \to \Theta^{B}$ (see  Fig.~\ref{fig: four-layer-nn-path}).
The output of the network at any intermediate point $\Theta_k^{AB}$ is approximately equal to $\widehat{\mathbf{Y}}^A$ as we have appropriately adjusted the weights $W_k^{AB}$ in the layer $k$.

To connect $\Theta_n^{AB} \to \Theta^{B}$ we need to change only the last layer. We can use any of our methods to do so. Note that it is sufficient to use the simple linear interpolation if the loss function is convex with respect to the last layer.

To connect any intermediate points $\Theta_k^{AB} \to  \Theta_{k+1}^{AB}$ or $\Theta^{A} \to  \Theta_{2}^{AB}$ note that $\Theta_k^{AB}$ and $\Theta_{k+1}^{AB}$ differ only in layers $k$ and $k+1$. So we can consider these two layers in  $\Theta_k^{AB}$ and $\Theta_{k+1}^{AB}$ as One Hidden layer subnetworks. The inputs of these subnetworks are identical, and the outputs are approximately the same thanks to the weight adjustment.  This means we can use any method we describe in Section \ref{sec:shallow} to connect the weights of these subnetworks.

The name convention is similar to the one we use for One Hidden layer network. We refer to the method as \textbf{Linear + Butterfly}, \textbf{Arc + Butterfly} or \textbf{OT + Butterfly} if we connect One Hidden layer subnetworks of intermediate points using the Butterfly weight representation and one of our distributional method. Alternatively, in the methods \textbf{Linear + Weight Adjustment}, \textbf{Arc  + Weight Adjustment} and \textbf{OT + Weight Adjustment} we consider the rows in the weight matrix of the first subnetwork layer as samples, connect them with one of our methods, and  perform weight adjustment on the second layer.

Let us also note that in case of  \textbf{Butterfly}--methods we can skip the $\Theta_n^{AB} $ intermediate point from the proposed path, so it becomes $\Theta^A \to \Theta_2^{AB} \to  \Theta_3^{AB} \to ,..., \to \Theta_{n-1}^{AB} \to \Theta^{B}$.

\subsection{CNNs and networks with skip connections}
The connection methods described above can be naturally generalized to convnets. In this case, the analog of the distribution of neurons would be the distribution of filters (since different filters can be viewed as independent, permutable entities). Of course, the distributional point of view should be more efficient if the number of filters is large. Our experiments below include connection of convnets such as VGG16. 

In the present paper we do not consider connection for networks with skip connections such as ResNets, mainly because the implementation in this case is relatively complex. We remark, however, that it is rather clear how that can be done by generalizing the stepwise procedure of Section \ref{sec:multi}: proceed layer-by-layer; in each layer, connect directly the weights sitting on all the incoming edges from earlier layers, and then adjust accordingly all the outgoing edges. 

\section{Experiments}
\label{sec: Experiments}
In this section, we test experimentally the proposed connection methods on the datasets CIFAR10 and MNIST. For each method, we measure the worst accuracy that the method provides along the path. We both datasets, we use the standard train--test split.

All considered models were trained using the cross-entropy loss with the SGD optimizer, for 400 epochs and 30 epochs on CIFAR10 and MNIST, respectively, with learning rate $0.01$ and batch size 128. For CIFAR10  we use the same standard data augmentation as \cite{huang2017snapshot}. For MNIST we do not use any augmentation. The activation function in all the networks is ReLU.

We compare our methods with connection curves numerically found in \cite{garipov2018loss}. In Table \ref{table: accuracy_connection_results}, \textbf{Garipov (3)} refers to the polygon with two 
segments, \textbf{Garipov (5)} refers to the polygon with four segments between the end points. Each Garipov's curve was optimized for 200 and 60 epochs for CIFAR10 and MNIST datasets, respectively, with batch size 128 as described in the original paper.

\subsection{One Hidden Layer}
Table \ref{table: accuracy_connection_results} shows results    for One Hidden Layer networks with 2000 hidden neurons, on MNIST and CIFAR10. 

As explained in Section \ref{sec: learnable_methods}, learnable methods (IAF flow, RealNVP bijection) require us to first collect a set $V$ of low-loss weight vectors $\Theta$, to be used for learning the connection methods. We created such a set of 16 models using the same training procedure but different random weight initializations and dataset augmentations. 

The ``train'' and ``test'' columns in Table~\ref{table: accuracy_connection_results} refer to the respective subsets of MNIST and CIFAR10. In the case of learnable connection methods, learning only used the training part of the  dataset; moreover, both ``train'' and ``test'' results were computed for endpoints $\Theta^A, \Theta^B$ not belonging to the model set $V$ used to learn the connection method.  IAF Flow failed to converge on CIFAR10. In the methods involving Weight Adjustment (Section \ref{sec:weightadjust}), adjustment of the second layer was also performed using only the training part of the dataset. 

In the supplementary materials (Section \ref{sec:a}) we analyze
how the considered methods perform for other network widths.
    
\begin{table*}[h]
\caption{Train and test accuracy (\%) of different methods for networks with a single hidden layer. End Point values show accuracy at the ends of the path. WA is short for Weight Adjustment. We show mean and one standard deviation of the worst point along the path.
}
\label{table: accuracy_connection_results}
\centering
\begin{tabular}{ |p{3.cm}||p{2.1cm}|p{2.1cm}|p{2.1cm}|p{2.1cm}|}
 \hline
\multicolumn{1}{l}{} & \multicolumn{2}{c}{MNIST} & \multicolumn{2}{c}{CIFAR10} \\
 \hline
Methods & train & test & train & test \\
 \hline
 Linear    & $96.54 \pm 0.40$    &$95.87 \pm 0.40$ & $32.09 \pm 1.33$    & $39.34 \pm 1.52$ \\
 Arc & $97.89 \pm 0.11$  & $97.03 \pm 0.14$ & $49.97 \pm 0.86$    &$41.34 \pm 1.39$\\

 \hline
  IAF flow & $96.34 \pm 0.54$  & $95.80 \pm 0.45$ & $-$    & $-$\\
 RealNVP bijection & $98.50 \pm 0.09$  & $97.53 \pm 0.11$ & $63.46 \pm 0.27$    & $53.94 \pm 0.95$\\
 \hline
 Linear  + WA&  $98.76 \pm 0.01$  & $97.86 \pm 0.05$ & $52.63 \pm 0.59$  & $57.66 \pm 0.26 $\\
 Arc  + WA& $98.75 \pm 0.01$  & $97.86 \pm 0.05$ & $58.77\pm 0.32$  & $57.88 \pm 0.24$\\
  \hline
  OT & $98.78 \pm 0.01$ & $97.87 \pm 0.04 $ & $66.19 \pm 0.23$ & $56.49 \pm 0.46$\\
OT + WA & $98.92 \pm 0.01$ & $97.91 \pm 0.03$ & $67.02 \pm 0.12 $& $58.96 \pm 0.21$\\
 \hline
 Garipov (3) & $99.10 \pm 0.01$& $97.98 \pm 0.02$ & $68.51 \pm 0.08$& $58.74 \pm 0.23$ \\
  Garipov (5) & $99.03 \pm 0.01$& $97.93 \pm 0.02$& $67.20 \pm 0.12$& $57.88 \pm 0.32$ \\
 \hline
  \hline
End Points & $99.14 \pm 0.01$ & $98.01 \pm 0.03$ & $70.60 \pm 0.12$ & $59.12 \pm 0.26$ \\
 \hline
\end{tabular}
\end{table*}

\subsection{Three Layer Network}
In the column FC3 of Table \ref{table: ml_conv_connection_results} we report results for three-layer networks with $6144$ neurons in the first hidden layer and $2000$ neurons in the second hidden layer. 

In this table, \textbf{Linear} and \textbf{Arc} are the simplest baseline methods that do not involve any layer-wise stages. 
We  simply simultaneously connect the respective rows of the weight matrices ($W_1, W_2, W_3$, in the case of three-layer networks) independently of each other, by considering these rows as sampled from some distributions ( $p_1, p_2, p_3$, respectively), and using either  linear segments as in Eq.(\ref{eq:linconnect}) or arcs as in Eq.(\ref{normal_transform}) for connection.

In the supplementary materials (Section \ref{sec:a}) we show how
performance of the methods changes as we vary network depth. 

\begin{figure}[h]
\begin{minipage}{.5\linewidth}
\centering
\includegraphics[width=1.\linewidth]{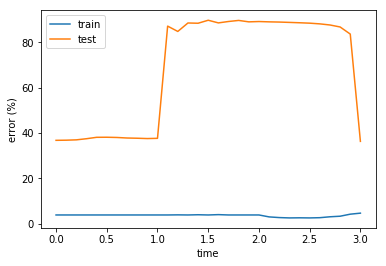}
 \end{minipage}%
 \begin{minipage}{.5\linewidth}
\centering
\includegraphics[width=1.\linewidth]{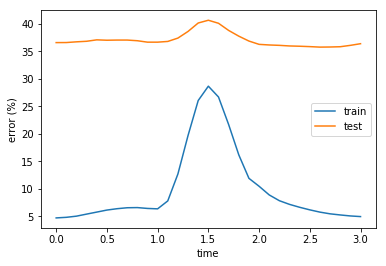}
 \end{minipage}
 \caption{Train and test error rates on a \textbf{Arc + Weight Adjustment} path connecting two local minima of a three-layer network. The intervals $[0,1],[1,2],[2,3]$ correspond to sub-paths $\Theta^A \to \Theta_2^{AB}, \Theta_2^{AB} \to \Theta_3^{AB},\Theta_3^{AB} \to \Theta^{B},$ respectively. 
 \textbf{Left:}  The train dataset is reduced to have a small size equal to the minimum hidden layer width of the network, $2000$. The reduced dataset is used both to perform Weight Adjustment and measure the accuracy of the method. 
 \textbf{Right:} Results with the full train dataset. 
 }
 \label{overparam_full_3layer_WA}
\end{figure}

Note that we make weight adjustment using the train dataset and report performance on the train and test datasets. However, it is clear from Table \ref{table: ml_conv_connection_results} that we do not observe the identical output along the path even on the train dataset. 
Let us point out why this can happen. 
Denote $\phi(W_2(0.5) \mathbf X^B_1)$ 
by $\mathbf X^{AB}_2$. 
Then the output of the network  
at the worst point of the connecting path on the dataset $\mathbf{X}$ is 
$\widehat{\mathbf{Y}}^A 
\approx W_3(0.5)\phi(W_2(0.5)\phi(W^B_1\mathbf {X})) 
= W_3^A  \mathbf {X}^{A}_2 
\big { [\mathbf {X}^{AB}_2 ] }^+ 
 \mathbf {X}^{AB}_2 $. 
The approximate equality becomes exact if  $\big {[ \mathbf{X}^{AB}_2]}^+ \mathbf{X}^{AB}_2 = I$. 
This happens when the network is overparameterized and all data points in ${\mathbf X^{AB}_2}$ are independent of each other, see the left side of Fig.~\ref{overparam_full_3layer_WA}. On the other hand, if we have more data points than neurons in the hidden layer, then  we only have the approximate inequality 
$\big { [\mathbf {X}^{AB}_2]}^+ \mathbf X^{AB}_2 \approx I$. Moreover, the more points we have compared to the number of neurons in the hidden layer, the more approximate this equality becomes. The underparameterized case is shown on the right side of Fig.~\ref{overparam_full_3layer_WA}. The drop of performance is clearly more drastic for the weight adjustment in the second hidden layer, which has only $2000$ hidden units (the interval $[1,2]$ in the plots), compared to the weight adjustment in the first hidden layer, which has $6144$ hidden units (the interval $[0,1]$).  

\subsection{Convolution Networks}
In columns Conv2FC1 and VGG16 of Table \ref{table: ml_conv_connection_results} we report results for the respective convolutional networks. Conv2FC1 is a simple network having $32$ and $64$ channels in the convolution layers (with kernel size $5$), and $3136$ neurons in the fully connected layer. VGG16 is used without batch normalization. 
The results show that methods without the WA procedure fail to construct low-loss paths for VGG16. Otherwise, the trends in the performance of different methods  are similar to those observed for dense multi-layer networks. See Section \ref{suppl: erd} in Supplementary materials for more results and discussion.

\begin{table*}[h]
\caption{The same as Table \ref{table: accuracy_connection_results} but for complex networks on CIFAR10. B-fly is short for Butterfly.
}
\label{table: ml_conv_connection_results}
\centering
\begin{tabular}{ |p{2.cm}||p{2.cm}|p{2.cm}||p{2.cm}|p{2.cm}|p{2.cm}|p{2.cm}|p{2.cm}|}
 \hline
\multicolumn{1}{l}{}  & \multicolumn{2}{c}{FC3} & \multicolumn{2}{c}{Conv2FC1} & \multicolumn{2}{c}{VGG16} \\
 \hline
Methods & train & test & train & test & train & test \\
 \hline
 Linear   &$31.10 \pm 0.84$ & $27.19 \pm 1.12$  &$25.86 \pm 4.62$ & $25.41 \pm 4.54$  & $10. \pm 0.$ & $10. \pm 0.$ \\
 Arc   & $ 46.39 \pm 1.03 $ & $40.17 \pm 0.84$ &$31.03 \pm  2.01$ & $30.44 \pm 2.09$  & $10. \pm 0.$ & $10. \pm 0.$ \\
  \hline
 Linear + B-fly  &$47.81 \pm 0.76$ & $38.38 \pm 0.84$  &$44.08 \pm 3.59$ & $42.46 \pm 3.43$  & $8.41 \pm 3.79$ & $8.57 \pm 3.49$ \\
 Arc + B-fly  &$ 60.60 \pm 0.79 $ & $49.63 \pm 0.86$  &$ 56.67 \pm 3.93 $ & $54.56 \pm 3.73$  & $3.67 \pm 4.56$ & $4.54 \pm 4.30$ \\
  \hline
 Linear + WA    & $60.93 \pm 0.25$    &$51.87 \pm 0.24$  & $71.09 \pm 0.38$    &$67.07 \pm 0.49$  & $94.16 \pm 0.38$    &$87.55 \pm 0.41$  \\
 Arc  + WA & $71.10 \pm 0.23$  & $58.86 \pm 0.29$ 
 & $77.36 \pm 0.99$  & $73.77 \pm 0.88$ & $95.35 \pm 0.239$  & $88.56 \pm 0.28$ \\
  \hline
 OT + B-fly & $81.95 \pm 0.29$ & $59.11 \pm 0.46$ & $76.94 \pm 1.41$& $73.66 \pm 1.44$ & $75.42 \pm 18.83$ & $68.56 \pm 17.80$ \\
 OT + WA & $87.53 \pm 0.18$ & $61.67 \pm 0.49$ 
 & $82.37 \pm 0.44$& $78.11 \pm 0.61$ & $96.61 \pm 0.18$ & $89.24 \pm 0.14$ \\
  \hline
  Garipov (3) & $94.56 \pm 0.08$& $61.38 \pm 0.36$ & $85.10 \pm 0.25$& $80.95 \pm 0.16$ & $99.69 \pm 0.03$ & $91.25 \pm 0.14$ \\
 \hline
 \hline
End Points & $95.13 \pm 0.08$ & $63.25 \pm 0.36$ & $87.18 \pm 0.14$ & $82.61 \pm 0.18$ & $99.99 \pm 0.$ & $91.67 \pm 0.10$  \\
 \hline
\end{tabular}
\end{table*}

\section{Ensembling with Weight Adjustment}\label{sec:ensembling}

In \cite{izmailov2018averaging} the authors perform averaging of several neural networks lying near each other in the weight space. Such close networks are taken from those obtained by SGD iterations. \cite{izmailov2018averaging} show that this leads to a better generalization and that such averaging approximates  ensembling of close models in the first order of approximation. The averaging has computational benefit compared to the usual ensemble of $n$ models that requires $n$ times more computation. 

In this section we propose another method to perform ensembling via weight averaging,  
applicable to any finite set of models on the weight manifold (typically, models optimized with different randomly chosen initial weights). The method is based on the Weight Adjustment procedure described in Section \ref{sec:weightadjust}. 

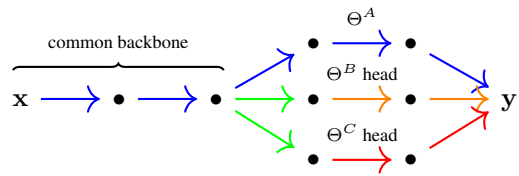
\begin{figure}
\begin{center}
\adjustbox{scale=1,center}{\hspace{-15mm}

\begin{tikzcd}[remember picture, row sep=small, column sep=0.8cm]
& & & & &
{\transparent{1}\bullet} 
\arrow[r, "\Theta^A" {black, yshift=1.5mm} 
opacity=1,
blue, thick] 
& 
{\transparent{1}\bullet} 
\arrow[dr, 
opacity=1, blue, thick] 
\\
& 
&
\mathbf{x}
\arrow[r, 
opacity=\opac, opacity=1,
blue, thick] 
& {\transparent{1}\bullet}  
\arrow[r, 
opacity=\opac, opacity=1,
blue, thick] 
& {\transparent{1}\bullet} 
\arrow[r, 
opacity=\opac, opacity=1,
green, thick] 
\arrow[ur, opacity=1, thick, blue]
\arrow[dr, 
opacity=1, thick, green] 
& {\transparent{1}\bullet} 
\arrow[r,"\Theta^B \text{ head}" {black, yshift=1.5mm}, , opacity=1, thick, orange] 
& {\transparent{1}\bullet} \arrow[r, , opacity=1, thick, orange] &
\mathbf{y}
\\
& & & & & 
{\transparent{1}\bullet}
\arrow[r, "\Theta^C \text{ head}" {black, yshift=1.5mm}, 
opacity=\opac, thick, red]
& 
{\transparent{1}\bullet}
\arrow[ur, 
opacity=\opac, red, thick] 
\end{tikzcd}
} 

\begin{tikzpicture}[overlay,remember picture]
\draw [
    thick,
    decoration={
        brace,
        raise=4.5mm
    },
    decorate
] (-3.3,1) -- (-0.5,1)
node [pos=0.5,anchor=north,yshift=10.5mm] {\scriptsize common backbone};
\end{tikzpicture}

\end{center}

\caption{A WA-ensemble of three models. Models $B$ and $C$ are adjusted to have the same backbone as model $A$. A longer common backbone reduces the amount of computation and required storage.
}
\label{fig: fish}
\end{figure}

The idea is to use the first network as a common backbone to extract features on some intermediate layer (see Fig.~\ref{fig: fish}). Given a particular data set and the $k$'th model, let us denote the output of this intermediate layer by $\mathbf F_k$ and the weights of the next layer by $W_k$. Also, we will denote by $head_k$ the computation performed in the $k$'th model after the multiplication by $W_k$.
Performing weight adjustment on the next layer,  $W_k^{1} = W_k \mathbf F_k \mathbf F_1^{+} $,  we make adjusted weights to operate on the same "basis" $\mathbf F_1$, for every  model $k$. Note that a net with these adjusted weights approximates the output of the $k$'th network. So, the average of the adjusted prediction $\frac 1 n \sum_{k=1}^n head_k (W_k^{1} \mathbf F_1)$ approximates the true ensembling of the models, where $n$ is the number of models in the ensemble. 

Note that if we adjust the last layer, there is no $head_k$ subnetworks to compute and we can just average the adjusted weights in the last layer. Moreover, if the loss is convex with respect to the model output, the loss of thus averaged models does not exceed the largest of the single model losses.

 \begin{figure}[h]
\centering
\includegraphics[scale=0.6, clip, trim=5mm 0mm 12mm 5mm]{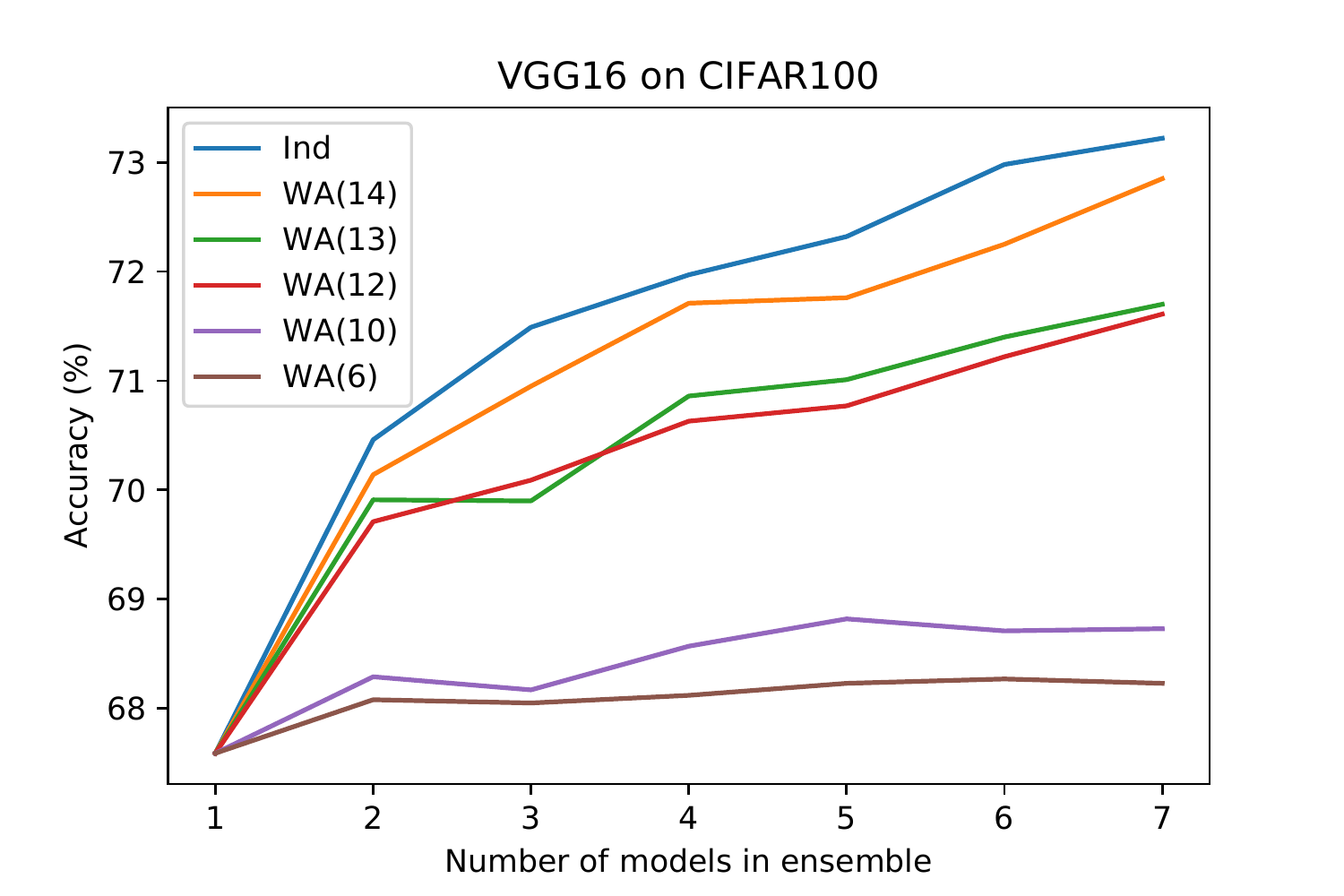}
\caption{Test accuracy (\%) of different WA-ensembles with respect to the number of models in the ensemble.
Ind corresponds to the ensemble of independent networks. WA(n) is WA-Ensemble with Weight Adjustment procedure performed on the $n$'th layer counting from the last network layer.}
 \label{WA_ensemples}
\end{figure}

In Figure \ref{WA_ensemples} we compare this WA-ensemble method against the usual ensemble of independently trained networks. We see that a longer common backbone reduces the amount of computation and required storage at the cost of accuracy:
the ensemble of independently trained models performs the best, followed by the WA(14) ensemble with two common layers,
 etc. We refer the reader to  Section \ref{sec: suppl ensembling} in supplementary material for more results.

\section{Discussion}
\label{sec: Conclusion}

We have described and compared a panel of generally applicable methods to connect a pair of weight vectors with a low-loss path. Our methods are  inspired by the distributional picture of weights in the networks and vary in complexity and accuracy. On the whole, our experiments show that on the realistic datasets such as MNIST and CIFAR10, our  connection methods are reasonably efficient, with efficiency naturally correlated with the complexity of the method.

The simplest nontrivial method -- Arc Connection -- is practically as simple and explicit as the baseline linear connection, but nevertheless provides a consistent improvement over the latter.
The learnable methods (IAF flow, ReLNVP bijection) further improve performance, thanks to taking into account the actual distribution of neurons. 

Optimal Transportantion and Weight Adjustment  perform even better, approximately matching and in some cases even slightly improving the direct numerical optimization results of \cite{garipov2018loss}. The key difference between them and the learnable methods is that the latter transform a neuron to the given state disregarding the states of the other neurons. In contrast, transformation of a single neuron under OT and WA takes into account the states of all neurons, which obviously creates an opportunity for a lower loss connection, at the cost of a higher computational complexity. 

The observed efficiency of the Optimal Transportation confirms the distribution-based explanation of the low-loss structure of the loss surface. Note, however, that the path constructed by OT is a rather complex piecewise linear curve, with the number of pieces scaling linearly with the network size. Also, this construction depends on the initial neuron matching that requires a separate optimization for each pair of endpoints. In contrast, global learnable methods (IAF flow, RealNVP bijection), while not achieving the accuracy of OT, provide relatively simple paths that depend on the endpoints only through the explicit arc formula.  

Summarizing, our results provide a relatively clear picture of connectedness of local minima in a large network. We see that natural ``macroscopic'' ideas lead to relatively simple low-lying paths, which can be further improved by taking into account more ``microscopic'' details. The resulting connection performance agrees with  previously known experimental results. Moreover, we have shown that low-loss connection paths give rise to a new kind of ensembling capable of improving the accuracy of the trained model with only a moderate increase of its complexity. It would be interesting to further explore the structure of connecting paths, 
with the view of a further computational simplification and better guarantees of performance improvement.  

\section{Acknowledgment}
IA thanks Denis Korzhenkov for helpful discussions and comments. DY acknowledges support of Huawei in the framework of the joint Skoltech--Huawei project ``CNN expressiveness''. 

\bibliography{paper}
\bibliographystyle{iclr2020_conference}

\appendix



\section{Experiments with different architectures}\label{sec:a}

We made a few additional experiments to see how the considered methods perform on different architectures.
In particular, we observe how the considered methods perform while we vary 1) the width of One Hidden layer network (Table \ref{table: width_test}) and 2) the depth of a dense network (Table \ref{table: depth_test}). All experiments were done on CIFAR10. 

In  Table \ref{table: width_test}  we  present experiments with underparameterized as well as overparameterized One Hidden layer networks. For any number of parameters, we observe approximately the same pattern of dependence of connection quality on the connection method.  Performance of most connection methods degrades at smaller numbers of parameters, but this is to be expected from the general logic of the distributional approach. 

in Table \ref{table: depth_test} we consider networks with 3, 5 or 7 layers.  We use 6144 neurons in the first hidden layer, 2000 neurons in the second hidden layers, and 1000 neurons in each of the remaining layers. Perhaps the most interesting observation that one can make here is that increasing depth from 3 to 5 improves performance of almost all connection methods. 

To connect minima with Garipov's curves we use the original implementation of their numerical algorithm (\url{https://github.com/timgaripov/dnn-mode-connectivity.git}).

\begin{table*}[h]
\caption{Test accuracy (\%) of different methods for One Hidden layer networks with different width on CIFAR10.}
\label{table: width_test}
\centering
\begin{tabular}{ |p{4.5cm}||p{2.1cm}|p{2.1cm}|p{2.1cm}|p{2.1cm}|}
 \hline
\multicolumn{1}{l}{} & \multicolumn{4}{c}{Width} \\
 \hline
Methods & 100 & 500 & 1000 & 2000 \\
 \hline
 Linear    & $33.20 \pm 2.03$    &$35.39 \pm 1.42$ & $36.35 \pm 1.68$    & $39.34 \pm 1.52$ \\
 Arc & $35.82 \pm 1.64$  & $36.73 \pm 1.44$ & $38.07 \pm 1.41$    &$41.34 \pm 1.39$\\
 \hline
 Linear  + Weight Adjustment&  $45.89 \pm 0.54$  & $53.56 \pm 0.33$ & $55.55 \pm 0.19$  & $57.66 \pm 0.26$\\
 Arc  + Weight Adjustment& $46.13 \pm 0.46$  & $53.84 \pm 0.32$ & $55.82 \pm 0.19$  & $57.88 \pm 0.24$\\
 \hline
   OT   & $53.73 \pm 0.41$  & $56.86 \pm 0.40$ & $56.18 \pm 0.18$  & $56.49 \pm 0.46 $\\
 OT + Weight Adjustment    & $55.10 \pm 0.35$  & $59.04 \pm 0.17$ & $58.95 \pm 0.19$  & $58.96 \pm 0.21 $\\
  \hline
  Garipov (3)  & $53.94 \pm 0.35$  & $58.47 \pm 0.21$ & $58.99 \pm 0.16$  & $58.74 \pm 0.23$\\
Garipov (5)  & $53.81 \pm 0.35$  & $57.59 \pm 0.27$ & $57.89 \pm 0.25$  & $57.88 \pm 0.32$\\
 \hline
  \hline
End Points & $56.47 \pm 0.26$ & $59.51 \pm 0.32$ & $59.14 \pm 0.23$ & $59.12 \pm 0.26$ \\
 \hline

\end{tabular}
\end{table*}

\begin{table*}[h]
\caption{Test accuracy (\%) of different methods for Dense networks with different depths on CIFAR10.}
\label{table: depth_test}
\centering
\begin{tabular}{ |p{4.5cm}||p{2.1cm}|p{2.1cm}|p{2.1cm}|p{2.1cm}|}
 \hline
\multicolumn{1}{l}{} & \multicolumn{3}{c}{Depth} \\
 \hline
Methods & 3 & 5 & 7  \\
 \hline
 Linear    & $27.19 \pm 1.12$    &$30.70 \pm 2.21$ & $25.43 \pm 2.04$   \\
 Arc & $40.17 \pm 0.84$  & $37.92 \pm 1.84$ & $35.94 \pm 2.77$   \\
 \hline
 Linear  + Butterfly&  $38.38 \pm 0.84$  & $50.66 \pm 0.83$ & $47.23 \pm 1.10$   \\
 Arc  + Butterfly & $49.63 \pm 0.86$  & $52.44 \pm 4.42$ & $47.47 \pm 3.31$   \\
  \hline
 Linear  + Weight Adjustment&  $51.87 \pm 0.24$  & $59.62 \pm 0.13$ & $58.12 \pm 0.16$   \\
 Arc  + Weight Adjustment& $58.86 \pm 0.29$  & $61.03 \pm 0.17$ & $60.15 \pm 0.14$  \\
 \hline
  OT + Butterfly  & $59.11 \pm 0.46$  & $60.78 \pm 0.39$ & $59.89 \pm 0.44$  \\
  OT + Weight Adjustment    & $61.67 \pm 0.49$ & $61.29 \pm 0.21$ & $60.35 \pm 0.24$ \\
   \hline
  Garipov(3)  & $61.38 \pm 0.36$  & $60.42 \pm 0.19$ & $58.95 \pm 0.18$ \\
  Garipov(5)  & $60.75 \pm 0.32$  & $59.51 \pm 0.21$ & $58.02 \pm 0.27$ \\
 \hline
  \hline
 End Points & $63.25 \pm 0.36$ & $61.72 \pm 0.21$ & $61.02 \pm 0.24$  \\
 \hline
\end{tabular}
\end{table*}

\section{Ensembling with Weight Adjustment}
\label{sec: suppl ensembling}

In Table \ref{table: ensemble_test}  we compare WA ensemble methods against ensembles of independently trained networks. WA($n$) in the table refers to Weight Adjusment procedure that is performed on the $n$'th layer counting from the last layer of neural network (e.g. WA(1) is an ensemble with the last layer adjusted).  We can see from the table that the amount of diversity in the ensemble is crucial for the performance: the more diversity (i.e., the higher $n$), the more accurate the output is. However, it comes with a cost of additional computations on inference and required storage. Also, note that the method WA(1)  slightly improves the results over one model, and it comes without the costs listed above. 

These results were obtained for VGG16 \cite{simonyan2014very} and PreResNet110 \cite{he2016deep} trained with SGD for 400 epochs, with learning rate 0.01 and batch size 128. We use standard data augmentation  as in \cite{huang2017snapshot}. We train VGG16 without batch normalization. 

We use the following implementations of VGG16 and PreResNet110.
\begin{itemize}
    \item VGG16: \url{https://github.com/pytorch/vision/blob/master/torchvision/
models/vgg.py}
\item PreResNet110: \url{https://github.com/bearpaw/
pytorch-classification/blob/master/models/cifar/preresnet.py}
\end{itemize}

\begin{table*}[h]
\caption{Test accuracy (\%) of ensemble methods with respect to number of models in ensemble and architectures on CIFAR10. }
\label{table: ensemble_test}
\centering
\begin{tabular}{ |p{2.5cm}||p{1.5cm}|p{2.1cm}|p{2.1cm}|p{2.1cm}|p{2.1cm}|}
 \hline
\multicolumn{1}{l}{} & \multicolumn{5}{c}{Number of models in ensemble} \\
 \hline
Architecture & method & 1 & 3 & 5 & 7  \\
 \hline
 \multirow{3}{*}{Dense 3}   
 & $WA(1)$    &$63.12$ & $64.22$  & $64.53$ & $64.53$  \\
 & $WA(2)$    &$63.12$ & $65.35$  & $66.27$ & $66.69$  \\
 & $Ind$    &$63.12$ & $65.67$  & $66.6$ & $67.04$  \\
  \hline
  \multirow{5}{*}{Dense 5}  
   & $WA(1)$    &$61.71$ & $62.44$  & $62.51$ & $62.68$  \\
    & $WA(2)$    &$61.71$ & $62.48$  & $62.77$ & $62.82$  \\
    & $WA(3)$    &$61.71$ & $62.99$  & $63.42$ & $63.64$  \\
    & $WA(4)$    &$61.71$ & $63.1$  & $63.71$ & $64.19$  \\
    & $Ind$    &$61.71$ & $63.07$  & $63.8$ & $64.33$  \\
 \hline
  \multirow{6}{*}{Dense 7}   
  & $WA(1)$    &$60.81$ & $61.13$  & $61.23$ & $61.2$ \\
  & $WA(2)$    &$60.81$ & $61.54$  & $61.74$ & $61.89$ \\
& $WA(3)$    &$60.81$ & $62.25$  & $62.53$ & $62.64$ \\
& $WA(4)$    &$60.81$ & $63. $  & $63.35$ & $63.6$ \\
& $WA(5)$    &$60.81$ & $62.97$  & $63.46$ & $63.63$ \\
& $Ind$    &$60.81$ & $63.35$  & $63.78$ & $64.01$  \\
 \hline
 \multirow{3}{*}{VGG16}   
 &   $WA(1)$    &$91.52$ & $91.54$  & $91.58$ & $91.59$   \\
  &   $WA(2)$    &$91.52$ & $91.64$  & $91.62$ & $91.61$   \\
  & $Ind$    &$91.52$ & $92.88$  & $93.12$ & $93.4$   \\
 \hline
  \multirow{2}{*}{PreResNet110} 
 &   $WA(1)$    &$92.49$ & $92.46$  & $92.53$ & $-$   \\
 &   $Ind$    &$92.49$ & $93.81$  & $94.08$ & $-$   \\
 \hline
\end{tabular}
\end{table*}

\section{Error rate dynamics along the path }
\label{suppl: erd}

\begin{figure}[tbh]
    \centering
    \includegraphics[width=\linewidth]{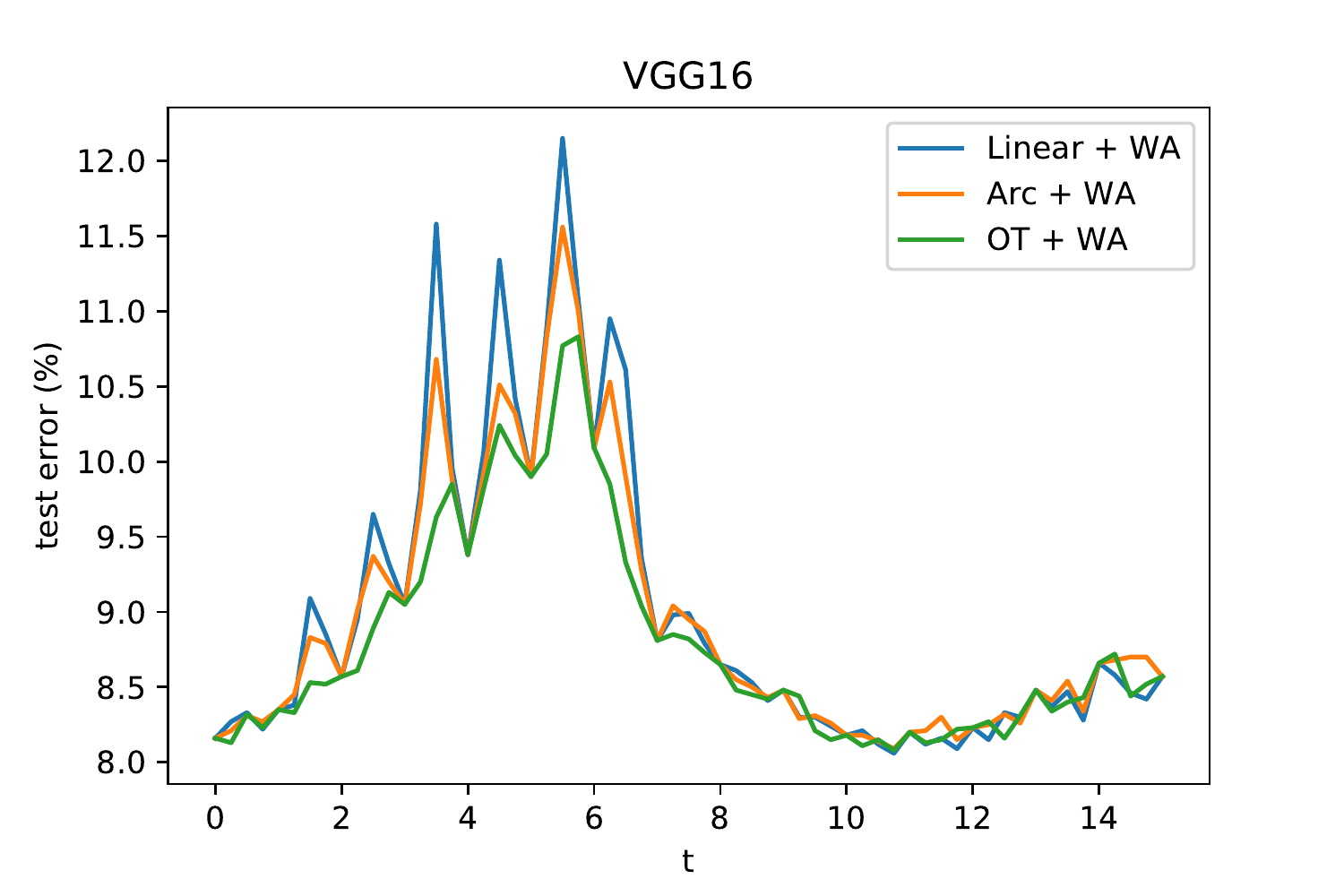}
    \caption{Test error of WA-based methods along the path on CIFAR10.}
    \label{fig: WA-based}
  \end{figure}

In Fig. \ref{fig: WA-based}  we show how test error changes along the paths proposed by WA-based connection methods for VGG16 on CIFAR10 dataset. The observed oscillations are associated with the 15 intermediate layer-by-layer transitions.

    \begin{figure}[tbh]
    \centering
    \includegraphics[width=\linewidth]{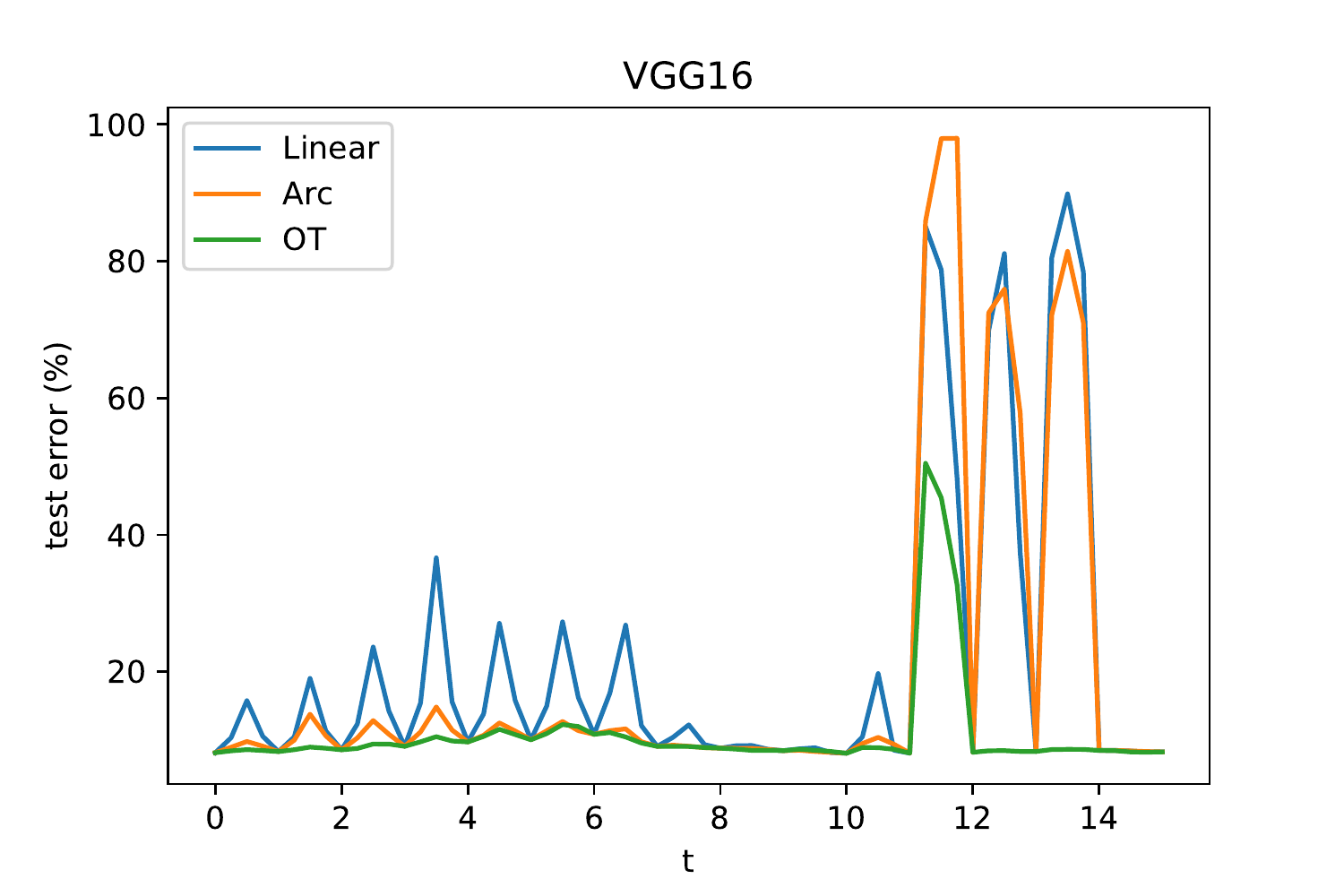}
    \caption{Test error of Butterfly methods along the connecting path (on CIFAR10).}
    \label{fig: Bfly-based}
  \end{figure}

In Table 2 of the main text we see that some connection methods fail to connect two minima of VGG16 network. Namely, Linear, Arc, Linear + Butterfly and Arc + Butterfly has accuracy equals to random guess or even lower. OT + Butterfly method performs better, but has a high variance -- we are currently investigating this issue. Note that neither of these methods uses Weight Adjustment procedure to improve the results.  In Fig.~\ref{fig: Bfly-based} we show examples of paths with failed Butterfly connections. As we can see, all Butterfly methods have low connection errors up to the 12’th layer, after which the errors increase drastically.

\begin{figure}[h]
    \centering
    \includegraphics[width=\linewidth]{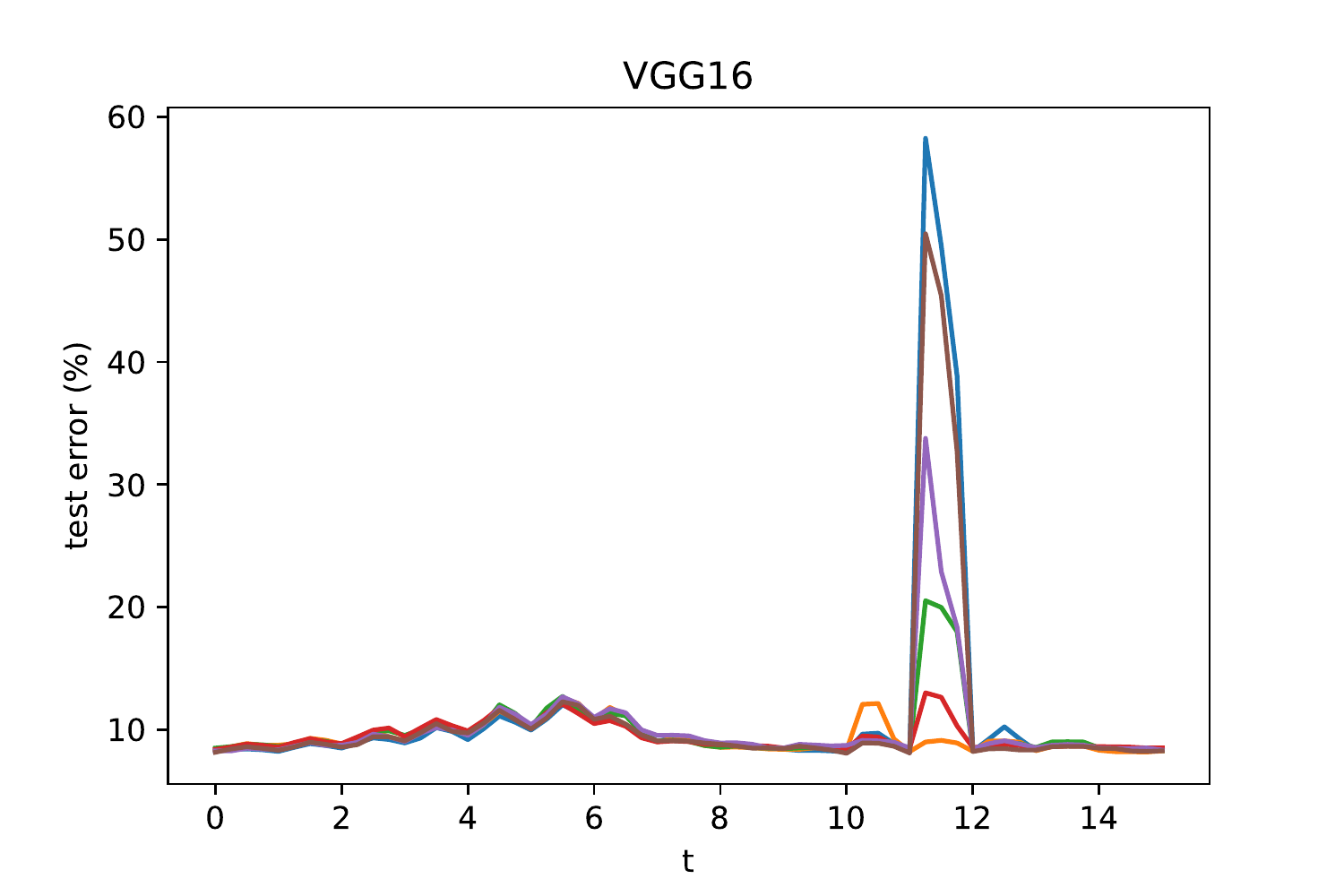}
    \caption{Test error of the method OT along six different paths (on CIFAR10).}
    \label{fig: OT}
  \end{figure}

Finally, we illustrate the variance of the OT + Butterfly method. In Fig.~\ref{fig: OT} we show accuracy on six different paths and observe that the main variance happens again on the  12'th layer.
 
\end{document}